\pdfoutput=1
\documentclass[11pt]{article}

\usepackage[]{acl}

\usepackage{times}
\usepackage{latexsym}
\usepackage{booktabs}

\usepackage[T1]{fontenc}
\usepackage[utf8]{inputenc}
\usepackage{pgfplots,pgfplotstable}
\usepackage{microtype}
\usepackage{cleveref}
\usepackage[colorinlistoftodos]{todonotes}

\title{HERDPhobia: A Dataset for Hate Speech against Fulani in Nigeria}

\author{Saminu Mohammad Aliyu$^{1,2,3}$, Gregory Maksha Wajiga$^{2}$, Muhammad Murtala$^{2}$,\\
\textbf{Shamsuddeen Hassan Muhammad$^{1,3,4}$, Idris Abdulmumin$^{3,5}$, Ibrahim Said Ahmad$^{1,3}$\medskip} \\
$^{1}$Bayero University, Kano, Nigeria, $^{2}$Modibbo Adama University, Yola, $^{3}$HausaNLP, \\
$^{4}$LIAAD-INESC TEC, Porto, Portugal, $^{5}$Ahmadu Bello University, Zaria, Nigeria\medskip \\
\footnotesize 
\texttt{smaliyu.cs@buk.edu.ng} %gwajiga@mautech.edu.ng, mmurtala1875@mautech.edu.ng,}\\
}

\begin{document}
\maketitle
\begin{abstract}

%The rate at which people are spreading hate speech against Fulani herders in Nigeria is alarming.  All crimes are blamed on them, not minding whether they are involved or not. Therefore, this paper presents the first corpus developed to automatically detect hate speech against Fulani herdsmen. It

%\todo[inline, color=blue!50]{Wannan emails din , ai na saminu kawai is ok. 
%Idris, what d u think?}%
%\todo[inline, color=red!50]{Is the table provided %(\Cref{tab:hate_cases}) sufficient? }%

Social media platforms allow users to freely share their opinions about issues or anything they feel like. However, they also make it easier to spread hate and abusive content. The Fulani ethnic group has been the victim of this unfortunate phenomenon. This paper introduces the HERDPhobia---the first  %Twitter 
annotated hate speech dataset on Fulani herders in Nigeria---in three languages: English, Nigerian-Pidgin, and Hausa.
%e describe the data collection method, the annotation process, and the baseline experiment. 
We present a benchmark experiment using pre-trained languages models to classify the tweets as either hateful or non-hateful. Our experiment shows that the XML-T model provides better performance with 99.83\% weighted F1.   
%speech or not hate speech, and the results were promising. 
We released the dataset\footnote{https://github.com/hausanlp/HERDPhobia} for further research.
\end{abstract}

\section{Introduction}

%A huge amount of data is being generated daily on the Web %Most of the data comes 
%from social media platforms such as Facebook, Twitter, YouTube, and Instagram. %Although people use these platforms to share good content, others
%Users take advantage of the anonymity provided by these platforms to post hateful and offensive content \cite{fortuna2018survey, djuric2015hate}. %This kind of post could
%that have a negative impact on the target and in some cases lead to suicide \cite{plaza2021comparing}. 

Hate speech is any form of communication that promotes hatred against an individual or a group based on religion, sexual orientation, race, ethnicity, gender, nationality, age, and disability \cite{schmidt2017survey}. %Other related terms used to refer to hate speech include cyberbullying, offensive, toxic, aggressive, abusive, and profane language \cite{del2021offendes}. Hate speech among Nigerians is mainly about ethnicity or religion. The task of regulating hate speech depends both on the government and on online platforms. As a result, platforms such as Facebook, Twitter, and YouTube designed policies that prohibit the posting of hateful or offensive contents \cite{warner2012detecting}.
Recently, the Natural Language Processing (NLP) community has seen an increase in research related to the task of detecting hate speech \cite{alkomah2022literature,poletto2021resources, celli2021policycorpus, moy2021hate}. 

%This is evident in many survey works\cite{alkomah2022literature, poletto2021resources, fortuna2018survey, fortuna2018survey, schmidt2017survey}

%fortuna2018survey, fortuna2018survey, }, \cite{fortuna2018survey}, \cite{alrehili2019automatic}, \cite{poletto2021resources} and \cite{alkomah2022literature}.% While some works focused on techniques for the detection of hate speech such as \cite{djuric2015hate} and \cite{del2017hate}; others focused on detecting specific instances of hate, e.g., misogyny \cite{pamungkas2020misogyny,zeinert2021annotating}; homophobia \cite{chakravarthi2021dataset}; and hate against immigrants \cite{sanguinetti2018italian}.

The Fulani herdsmen in Nigeria are known %to be involved in pastoral activities. They 
for their migration, along with their herds from one location to another for grazing purposes \cite{enor2019contending}. %Some Fulani herdsmen are among several other criminals that attack and kidnap Nigerians \cite{ojo2020governing}.
%However, %since Muhammad Buhari emerged as the president of Nigeria, a Fulani man, 
Recently, there has been an exponential increase in hate rhetoric against the Fulani tribe \cite{udanor2019combating}. Hence, there is a need to develop an automatic system for the detection of hate speech. 
%Developing a good corpus is very critical to a successful automatic system for hate detection. 

In this paper, we present HERDPhobia, a Twitter dataset for hate speech detection against Fulani herdsmen. 
%We collect data from Twitter, as it is the most widely used online platform by Nigerians to express their emotions. 
To our knowledge, this is the first dataset created for the detection of hate speech against Fulani herdsmen in Nigeria.

%The remaining part of the paper is organised as follows: \cref{lit-rev} reviews related literature. In \cref{data}, we describe the process of data collection and annotation of the HERDPhobia corpus. \Cref{models} discusses baseline evaluation of the corpus. Finally, \cref{conclusion} discusses our conclusion and future work.

\section{Related work}
\label{lit-rev}
%Hate speech detection has attracted the attention of researchers in the Natural Language Processing community. Researchers proposed different techniques to detect hate speech \cite{moy2021hate}, as evident in the survey works by \cite{schmidt2017survey}, \cite{fortuna2018survey}, \cite{alkomah2022literature},  \cite{poletto2021resources} and \cite{alrehili2019automatic}.
%One crucial aspect of automatic hate speech detection systems is the dataset.
Some of the work done to create hate dataset for automatic detection includes: a corpus of 16,914 English tweets to detect sexist or racial slur by \citet{waseem2016hateful};
%The corpus contains a total of 16,914 tweets that were manually annotated by the researchers. The annotated tweets were review by a third annotator to reduce bias in the labelling. Their inter-annotator agreement score was 84 percent.
\citet{davidson2017automated} collected 85.5 million tweets from Twitter using a hate speech lexicon by Hatebase.org, manually annotating a random 25,000;
%tweets were randomly selected and manually annotated by CrowdFlower workers into “hate speech, offensive but not hate speech or neither offensive nor hate speech”. Most of the tweets were labelled as offensive with only 5 percent labelled as hate speech. They reported a very high inter-annotator agreement (92 percent).
\citet{poletto2017hate} created a corpus of hate against immigrants, Muslims, and Roma;
%They used keywords to crawl tweets in Italian language on the three classes and selected 700 tweets from each class, making a total of 2100 tweets. The corpus was annotated by four annotators and they reported a low Inter annotator agreement which they attribute to the complexity of the task.
\citet{ibrohim2018dataset} developed a dataset to detect abusive language in Indonesian tweets using a machine learning approach;
\citet{ousidhoum-etal-2019-multilingual} annotated multilingual hate corpus in English, French, and Arabic.
%They conducted two separate experiments; one as a binary classification (abusive and non-abusive language) and the other as a non-binary classification (abusive, offensive and offensive but not abusive language). Overall, NB was found to perform better than the other classifiers. 
and \citet{del2021offendes} created a corpus of comments from Twitter, YouTube, and Instagram for offensive language detection in Spanish.
%The comments were compared with five existing lexicons to determine the presence of offensive words and sorted according to their lexical diversity. The tweets were labelled by human annotators into five classes. A low inter-annotator score was reported and is attributed to the difficult of the task \cite{del2021offendes}.

\section{Data Collection and Annotation}
\label{data}

Following a similar approach for tweet collection used in
\cite{muhammad2022naijasenti}, we collected tweets using three keywords:\textit{~Fulani, cow and herdsmen} from Nigeria in three languages: English (97.2\%), Hausa (1.8\%) and Nigerian-Pidgin (1\%).

For data annotation, we adopted the guidelines from \cite{warner2012detecting}, and we annotated tweets into three categories: \textit{Hate (HT)}, \textit{Not-Hate (NHT)}, and \textit{Indeterminate(IND)}. A tweet was labelled as hate if it contains words that attack or disparage an individual or group that belongs to the Fulani tribe. All factual and non-sentimental tweets were classified as Not-Hate. Tweets that are sarcastic or ambiguous were classified as indeterminate. After the annotation, we obtained $1,131$ \textit{HT}, $5,007$ \textit{NHT}, and $36$ \textit{IND} tweets, with a Fleiss kappa inter-annotator agreement score of 57\%.

\begin{table*}[t]
    \scalebox{0.95}{
        \centering
        \begin{tabular}{lp{30em}p{2.5em}p{2.5em}p{2.5em}}
            \toprule
             S/N & Tweet & A1 & A2 & A3 \\ \midrule
             1 & @user it's true that all 
              \textcolor{blue}{
             fulani herdsmen} are 
             \textcolor{red}{terrorist} think wisely b4 answer & HT & HT & HT \\
             2 & @user how much for the whole.cow boss & NH & NH & NH\\
             %3 & @user granny squared bralette ft the ruffled short & IND & IND & IND\\
             3 & @user na so all for fulani malu & IND & HT & NH \\
            \bottomrule
        \end{tabular}
    }
    \caption{Hate Tweets Exmaples, A1 = Annotator1, A2 = Annotator2, A3 = annotator3}
    \label{tab:hate_cases}
\end{table*}

\begin{figure}[t]
    \begin{tikzpicture}
        \begin{axis}[
                ybar,
                symbolic x coords={Lagos,Abuja,Others,Anambra,Oyo,Delta,Ekiti,Osun,Ogun},
                xtick=data,
                xticklabel style={rotate=45},
                axis x line=bottom, 
                axis y line=left,
                axis line style={-},
                bar width=.2cm,
                enlarge x limits={abs=.5cm},
                nodes near coords
            ]
            \addplot[red!20!black,fill=red!80!white] table[x=location,y=count]{\statesStat};
        \end{axis}
    \end{tikzpicture}
\caption{Distribution of hateful tweets by states. 12 states have less than 10 tweets and are group as Others.}
\label{fig:states_stats}
\end{figure}
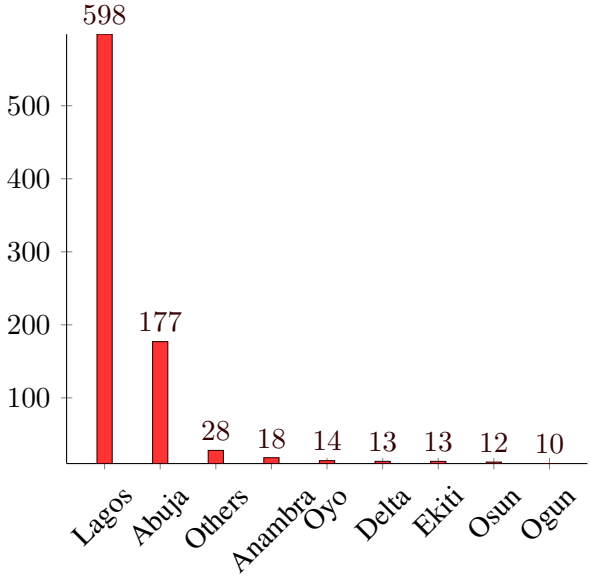

\iffalse 

\begin{table}[!ht]
\centering
\begin{tabular}{l r} 
 \toprule
 label & Number \\ \midrule
 Hate (HT) & 1,131  \\ 
 Not-hate (NHT) & 5,007  \\
 Indeterminate (IND) & 36 \\ \midrule
 Total & 6,174 \\ \bottomrule
\end{tabular}
\caption{Number of tweets in the three categories.}
\label{table:class}
\end{table}

\begin{description}
\item[Hate:] A tweet is labelled as hate if it contains words that attack or disparage an individual or group that belongs to the Fulani tribe. % For example, the tweet \textit{(“this cow is back jesus”)} is a Hate statement because the author is referring to the President of Nigeria, who is a Fulani man.
\item[Not-Hate:] All factual and non-sentimental tweets were classified as Not-Hate. %For example, although the word \textit{"fulani"} appeared in the code-mixed tweet: \textit{("wawu so cute dama nupe have a little linkage with fulani")}, it is not hate.
\item[Indeterminate:] Tweets that are proverbs, sarcasm or ambiguous  are classified as Indeterminate. 
See table 1 for examples of the classes of the tweets.
\end{description}

\fi 
%x were trained and provided with the annotation guidelines.
Three annotators annotated each tweet, and we used a simple majority vote to select the tweet to train our model from two classes (\textit{HT} and \textit{NHT}). 
%using only two classes (\textit{HT} and \textit{NHT}). 
\Cref{tab:hate_cases}
provides examples of annotated the tweets. The first two were straightforward cases were the three annotators unanimously agree on their labels as \textbf{Hate} and \textbf{Not Hate} respectively.
%as the tweeter labelled the Fulani herdsmen as terrorist. 
%Similarly, the second tweet was labelled \textbf{Not Hate} by all the annotators, as the usage of the word cow used in the sentence refers to the animal cow. %The third example was labelled \textbf{Indeterminate} as it does not convey any meaning. 
The last is a hard-case annotation scenario where the three annotators could not agree due to its ambiguity. \Cref{fig:states_stats} shows distribution of hateful tweets, with Lagos having the highest. Overall, hateful tweets are more prevalent in the southern part of Nigerian states, as also reported in \cite{udanor2019combating}.

\section{Baseline Model}
\label{models}
To prepare the training and validation data, we removed all tweets that are partially or completely annotated as indeterminate. This resulted in 892 Hate and 3,523 Not Hate tweets.
%Thereafter, we used a 70–10-20 percent split to create the Train, Dev and Test sets respectively. The final data statistics are provided
We splitted the dataset as shown in \Cref{table:train-data}.

\pgfplotstableread[row sep=\\,col sep=&]{
    location & count \\
    Lagos & 598 \\
    Abuja & 177 \\
    Others & 28 \\
    Anambra & 18 \\
    Oyo & 14 \\
    Delta & 13 \\
    Ekiti & 13 \\
    Osun & 12 \\
    Ogun & 10 \\
}\statesStat

\begin{table}[!ht]
\centering
\begin{tabular}{lrrr} 
 \toprule
 Data & Train & Dev & Test \\
 \midrule
 \# of tweets & 3,090 & 441 & 884 \\
 \bottomrule
\end{tabular}
\caption{Training and Validation Data Statistics.}
\label{table:train-data}
\end{table}

We trained three different models on the HERDPhobia dataset by fine-tuning 3 pretrained language models: mBERT, XLM-T and AfriBERTa to create a baseline. The models were trained using the same hyper-parameters as in \citet{shode_africanlp}: a total of 20 epochs using a batch size of 32 and a maximum sentence length of 128. The weighted f1-score obtained for each model on the test set is shown in \Cref{table:model-f1}. The best performance was obtained from XLM-T with 99.83\% weighted F1. This may be because HERDPhobia is dominated by English tweets, the same language that was used to pretrain XLM-T.

\begin{table}[t]
\centering
\begin{tabular}{lr} 
 \toprule
 Model & F1-score \\ 
 \midrule
 XLM-T \cite{barbieri-espinosaanke-camachocollados:2022:LREC} & 99.83  \\
 mBERT \cite{devlin-etal-2019-bert} & 80.96  \\ 
 AfriBERTa \cite{ogueji-etal-2021-small} & 78.07 \\
\bottomrule
\end{tabular}
\caption{Benchmark results with weighted F1 scores.}
\label{table:model-f1}
\end{table}

\section{Conclusion and Future Work}
While there are many forms of hate based on religious and ethnic stereotypes in Nigeria, the Fulani tribe suffers the most. In this paper, we present a new dataset for hate speech against Fulani. The dataset---HERDPhobia---consists of a total of 6,174 tweets that were manually annotated into three classes, hate, not-hate and indeterminate.
%{27\% percent tweets were labelled as hate speech 31\% percent as not-hate and 2\% percent} tweets as indeterminate. \textbf{To talk about the inter-annotator agreement score}.
Our baseline experiment with three pre-trained language models (XLM-T, mBert, AfriBerta) shows that XLM-T performs best with 99.83\% weighted F1.  
In future work, we plan to extend our work to detecting hate speech to more ethical groups in Nigeria. We also plan to include offensive language in the classification category and label hate tweets according to hate types and intensity.

\section{Ethical Consideration}
We replaced all personally identifiable information from the dataset. Mentions are replaced with \@user.

\label{conclusion}

%\subsection{Appendices}

%\section*{Acknowledgements}

%\subsection{References}
\bibliography{anthology,custom}
\bibliographystyle{acl_natbib}

%\appendix

%\section{Example Appendix}
%\label{sec:appendix}

%This is an appendix.

\end{document}